\documentclass[letterpaper, 10 pt, conference]{ieeeconf}  % Comment this line out if you need a4paper
\usepackage{graphicx}
\usepackage{amsmath}
\usepackage{bm}
\usepackage{subcaption}
\usepackage{tabularx,booktabs}
\usepackage{url}

\IEEEoverridecommandlockouts                              % This command is only needed if 
\title{\LARGE \bf
Collaborative motion planning for multi-manipulator systems through Reinforcement Learning and Dynamic Movement Primitives
}

\author{Siddharth Singh$^{*}$, Tian Xu$^{*}$ and Qing Chang$^{1}$% <-this % stops a space
\thanks{*Equal Contribution}% <-this % stops a space
\thanks{All the authors are with the Department of Mechanical \& Aerospace Engineering, University of Virginia,
        Charlottesville, VA 22903, USA}%
\thanks{$^{1}$Corresponding author
        {\tt\small qc9nq@virginia.edu}}%
}

\begin{document}

\maketitle
\thispagestyle{empty}
\pagestyle{empty}

%%%%%%%%%%%%%%%%%%%%%%%%%%%%%%%%%%%%%%%%%%%%%%%%%%%%%%%%%%%%%%%%%%%%%%%%%%%%%%%%
\begin{abstract}

Robotic tasks often require multiple manipulators to enhance task efficiency and speed, but this increases complexity in terms of collaboration, collision avoidance, and the expanded state-action space. To address these challenges, we propose a multi-level approach combining Reinforcement Learning (RL) and Dynamic Movement Primitives (DMP) to generate adaptive, real-time trajectories for new tasks in dynamic environments using a demonstration library. This method ensures collision-free trajectory generation and efficient collaborative motion planning. We validate the approach through experiments in the PyBullet simulation environment with UR5e robotic manipulators.
\textit{Project Website: 
\url{https://sites.google.com/virginia.edu/oncoldmp/home}}

\end{abstract}

%%%%%%%%%%%%%%%%%%%%%%%%%%%%%%%%%%%%%%%%%%%%%%%%%%%%%%%%%%%%%%%%%%%%%%%%%%%%%%%%
\section{INTRODUCTION}
% This paragraph is too long, we can use simply 1~2 sentences to state the important of the dual-arm system, and point out the disadavantages of current reasearch

% XT comment
% \textbf{Tian's version}\\
Compared to the single-arm robot system, multiple robots offer superior operation and control capabilities, particularly in coordinated tasks and human–machine collaboration \cite{dualarmsurvey}. 
% As industries increasingly adopt multi-robot systems, efficient and safety-aware motion planning methods are needed to enable online cooperative manipulation for multiple robots control. 
As industries increasingly adopt multi-robot systems, there is a critical need for advanced, safety-aware motion planning methods that can facilitate real-time cooperative manipulation, ensuring precise and efficient control of multiple robots in dynamic environments.

Currently, there are many motion planning methods available for robot arm control.
Based on the primary focus, these motion planning methods could be categorized into two key aspects: high-level task sequencing and low-level execution control.

At the high level, learning-based methods such as Imitation Learning (IL) (\cite{schaal, collab3}), Reinforcement Learning (RL) (\cite{RLSurvey, Levine}), and Graph Learning (\cite{Gl1, Gl2}) are frequently employed to sequence sub-tasks for each robotic arm based on the given task. While effective, these approaches often require extensive, costly datasets, limiting their scalability. Additionally, rule-based learning \cite{rulebased} and temporal logic \cite{logic1} are commonly used to decompose tasks into primitive motions, with high-level controllers producing a sequence of actions and inverse-kinematics solvers generating motion plans. However, these approaches primarily focus on task sequencing, with limited attention to motion planning and task execution integration.

At the low level, optimization techniques such as Model Predictive Control (MPC) (\cite{mpc1, mpc2}) and Trajectory Optimization \cite{to1} are used to compute optimal joint trajectories by leveraging dynamic models that prioritize safety and collision avoidance. Despite their precision, these methods are computationally intensive, making real-time implementation challenging. Alternatively, dynamic system-based methods like Dynamic Movement Primitives (DMP) have proven effective in generating stable, collision-free trajectories with minimal demonstration requirements \cite{nadia1, nadia2}. However, their imitation-driven nature limits their ability to enable higher-level collaboration among multiple robotic arms.
Ginesi et al. \cite{Ginesi2021, Ginesi8981552} proposed static and dynamic volume potential field methods that enable multiple robots to collaborate while avoiding self-collisions.
However, these methods often treat each arm as an obstacle, which hinders the generation of collaborative trajectories, particularly in novel scenarios.

% \begin{figure}
%     \centering
%     \includegraphics[width=1\linewidth]{First_complete.png}
%     \caption{Combining RL with enhanced DMP, the proposed method generates online trajectories in real-time to enhance collaboration.}
%     \label{fig:intro}
% \end{figure}
In this work, we introduce a method that leverages one-time human demonstrations to generate online executable trajectories for multi-arm robotic systems using Dynamic Movement Primitives (DMPs) to enable collaborative task completion. The proposed approach adopts a hierarchical structure. Given the task specifications, the higher level utilizes a library of human-demonstrated trajectories to independently generate a trajectory for each arm using Q-learning. These reference trajectories are then passed to the lower level, which manages online execution with a focus on collaboration and collision avoidance. To bolster generalizability, an optimization step is introduced which computes the parameters of both the DMP and the artificial potential field. Additionally, considering the end-effector pose, a new potential field calculation step is integrated. Finally, a heuristic approach is developed to enable real-time cooperation.  We refer to this enhanced DMP as (Optimized Normalized Collaborative) ONCol-DMP.

The main contributions of the proposed method are presented in two aspects:
\begin{itemize}
    \item \textbf{Integration of Kinematic Skill Learning and Dynamic Trajectory Planning}: Developing a unified framework that links kinematic skill learning with dynamic trajectory planning for effective real-world robotic execution.
    \item \textbf{Collaborative Execution Using ONCol-DMP and Heuristic Control}: Proposing a novel framework named Optimized Normalized Collaborative Dynamic Movement Primitives (ONCol-DMP) for efficient obstacle avoidance with a heuristic phase control technique to regulate execution speed, minimizing collisions and trajectory deviations, enabling seamless multi-robot collaboration.
\end{itemize}

\begin{figure*}[ht]
      \centering
      \includegraphics[width=\textwidth]{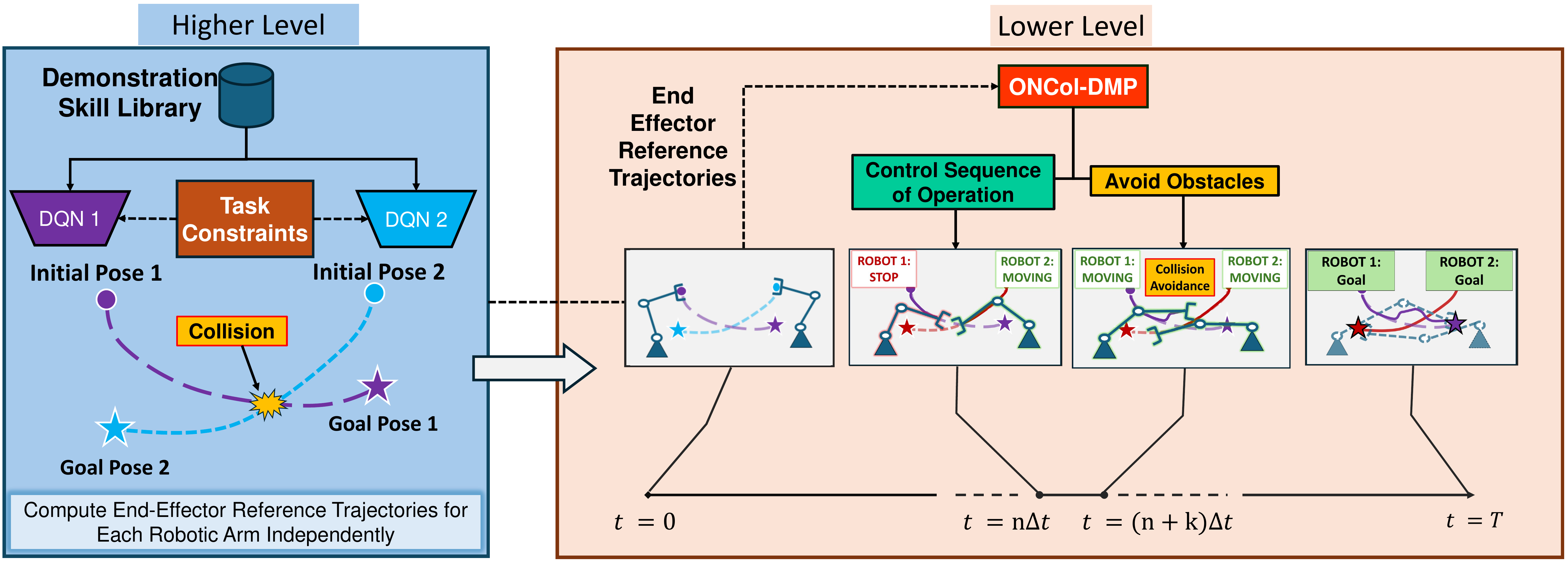}
      \caption{Overview of the proposed approach. The higher level, utilizing Q-Learning, generates independent motion plans. The proposed Collab-DMP ensures collision avoidance and can also control the sequence of the operation.}
      \label{fig:overview}
\end{figure*}

The proposed method is validated in a simulated environment across multiple tasks. Our results demonstrate that the method successfully generates real-time, collision-free trajectories for multiple robotic arms, allowing them to cooperatively complete the tasks. 

The rest of the paper is structured as follows: in Sec. \ref{sec:background} introduces the necessary mathematical background followed by problem formulation. Section \ref{sec:proposed_method} details our proposed method. The effectiveness and validation of our proposed contributions is presented in \ref{sec:exper}. Section \ref{sec:conclusion} entails the conclusion of our work and discusses the future work.

\section{BACKGROUND}\label{sec:background}
\subsection{Dual-quaternion based Featurization}
In this paper, we further extend the skill-learning method and adopt the featurization technique proposed in \cite{tianyu}.
For our purposes, the demonstrations are defined in SE(3) space to capture the pose of the end-effectors. We describe the poses using a dual quaternion encoding both the rotational and translational information. Given a $6-DoF$ pose $\boldsymbol{x} \in SE(3)$, the equivalent dual quaternion is defined as 

\begin{equation}\label{eq:dual_quatdef}
    \boldsymbol{q} = \boldsymbol{q}_r + \frac{1}{2}\eta(\boldsymbol{q}_t \otimes \boldsymbol{q}_r)
\end{equation}

where $\eta \neq 0$, but $\eta^2=0$ and $\otimes$ represents the quaternion multiplication. In eq. \eqref{eq:dual_quatdef}, $\boldsymbol{q}_t$ is the quaternion representing the pure translation of the rigid body, represented as 

\begin{equation}
    \boldsymbol{q}_t = (0, \hat{\boldsymbol{t}})
\end{equation}

where $ \hat{\boldsymbol{t}} = x \hat{\boldsymbol{i}} + y  \hat{\boldsymbol{j}} +  z \hat{\boldsymbol{k}}$ representing the translation in $SE(3)$. Similarly, in eq.\:\ref{eq:dual_quatdef}, $\boldsymbol{q}_r$ represents the rotational orientation of the rigid body which is defined as:

\begin{equation}
    \boldsymbol{q}_r = cos(\frac{\phi}{2}) + \hat{\boldsymbol{v}}sin(\frac{\phi}{2})
\end{equation}

where $\hat{\boldsymbol{v}} = v_x \hat{\boldsymbol{i}} + v_y  \hat{\boldsymbol{j}} +  v_z \hat{\boldsymbol{k}}$, is the unit vector in $SE(3)$ along the axis of rotation and $\phi$ is the angle of rotation. Given, a trajectory of the poses as dual quaternions, $\mathcal{T} = {\boldsymbol{x}^{ee}_0, \boldsymbol{x}^{ee}_1, \cdots, \boldsymbol{x}^{ee}_{N_p}}$, the featured trajectory is computed as $\Tilde{\mathcal{T}} = \{ \boldsymbol{\delta}_0, \boldsymbol{\delta}_1, \cdots, \boldsymbol{\delta}_{N_p - 1} \}$, where 
\begin{equation}
    \boldsymbol{\delta}_i = \boldsymbol{q}_i \otimes \boldsymbol{q}_{N_p}
\end{equation}

Additionally, to compare the similarity of any two poses we use the semantic similarity is defined as,
\begin{equation}\label{eq:semsim}
    S(\boldsymbol{\delta}_i,\boldsymbol{\delta}_k) = min\left(||\boldsymbol{\delta}_{ir} - \boldsymbol{\delta}_{kr}||, ||\boldsymbol{\delta}_{ir} + \boldsymbol{\delta}_{kr}|| \right)
\end{equation}

\subsection{Dynamic Movement Primitives}\label{sec:background_dmp}
Dynamic Motion Primitive (DMP) is a versatile framework for trajectory learning in robotics, based on an Ordinary Differential Equation (ODE) that models motion using a spring-mass-damper system with an added forcing term. We utilize discrete DMP, which is a linear, second order dynamic model with a nonlinear forcing term. We define DMP for a single DoF trajectory $x$ of a discrete movement is defined as follows:
\begin{equation}
 \tau \dot{z}=\alpha_z(\beta_z (g-x)-z)+f(s),
 \label{eq:trans1}
 \end{equation}
 \begin{equation}
     \tau \dot{x}=z,
     \label{eq:trans2}
 \end{equation}
\begin{equation}
     \tau \dot{s}=-\alpha_s s
     \label{eq:phase}
\end{equation}

 where $s$ is the phase variable and $z$ is an auxiliary variable. The damping parameters $\alpha_z$ and $\beta_z$ define the behavior of the second-order system. $\tau$ is a temporal parameter that defines the period of the trajectory. $\alpha_s$ is the parameter controlling the convergence speed of the phase variable $s$.

The Eqs.~\ref{eq:trans1} and \ref{eq:trans2} are called the transformation system, while the Eq. 8 is referred to as the canonical system.
$f(s)$ is defined as a linear combination of $C$ nonlinear Radial Basis Functions (RBFs), which enables the robot to follow any smooth trajectory:
\begin{equation}
f(s)=\frac{\sum^C_{i=1}w_i\Psi_i(s)}{\sum^N_{i=1}\Psi_i(s)s},
\end{equation}
\begin{equation}
    \Psi_i(s)=\mathrm{exp}\left({-h_i(s-c_i)^2}\right),
\end{equation}
where the weights $w_i$ could be updated by Locally Weighted Regression (LWR) \cite{Atkeson1997LocallyWL}.

\subsection{Problem Formulation}
We define the task specifications as the sequence end-effector poses $\mathcal{X}^{ee} := \{\boldsymbol{x}^{ee}_0, \boldsymbol{x}^{ee}_1, \cdots, \boldsymbol{x}^{ee}_{K}\}$, referred to as critical configurations. Here $K$ are the number of key configurations in the task trajectory. 
Given a task specifications as a sequence of end-effector poses for the $N$ robots in scene, $\mathcal{T} := \{\mathcal{X}^{ee1}, \mathcal{X}^{ee2}, \cdots, \mathcal{X}^{eeN} \}$, devise the joint-trajectory for each robot, $\{\boldsymbol{\Theta}^1, \boldsymbol{\Theta}^2, \cdots, \boldsymbol{\Theta}^N\}$ to follow the critical configurations of each robot end-effector.

\section{PROPOSED METHOD}\label{sec:proposed_method}

Figure \ref{fig:overview} shows  the overview of our proposed method. The hierarchical structure is introduced to separate the problem from preliminary trajectory generation from cooperation and collision avoidance. 

At the higher level, we first train a Deep Q-Network (DQN) agent, which, given the task specification for a robot, learns to generate a trajectory from the skill library. These DQN agents are then duplicated and provided to each manipulator. The trained DQN agent generates trajectories independently, i.e., without the awareness of the other manipulators present in the environment. These trajectories act as the end-effector reference trajectories for the DMPs at the lower level. At the lower level, each robot's individual DMP is responsible for executing the trajectory and ensuring collision avoidance. Additionally, the enhanced DMPs have a collaborative term that allows the DMPs to reduce the execution speed of the DMP, leading to exponentially-stable collision-free trajectories.

\subsection{Higher Level Trajectory Generation}\label{sec:high_level}
The problem of skill learning is posed as a Markov Decision Process (MDP). We define the state as a segment of task trajectory at hand, $s_t = \{\boldsymbol{x}^{ee}_t, ..., \boldsymbol{x}^{ee}_K\}$. The action is defined as a 2-tuple of the segment at hand and a corresponding demonstration trajectory allowing us to define the action space as $\mathcal{A}_t = \{ (s_t,\Tilde{\mathcal{T}}^d_i) \;|\; \forall \; i \; \in \{1, ..., N_d\} \}$, where $\boldsymbol{\delta}^d_i$ is the $i^{th}$ featurized demonstration and $N_d$ are total number of demonstrated skills. To identify the closest matching demonstration task, we compute the reward based on the semantic similarity (Eq.~\ref{eq:semsim}) of the two segments as: 
\begin{equation}
    r_t = \sum_{j=t}^K \sum_{l=1}^{N_{id}} S(\boldsymbol{\delta}^{ee}_j,\boldsymbol{\delta}^d_l)
\end{equation}

Here, $N_{id}$ are the number of critical-configurations in the $i^{th}$ demonstration.

We train a single DQN-agent and duplicate it for each robot individually. The task specification for the $r^{th}$ robot, i.e. $\mathcal{X}^{eer}$ is passed as the input to it's respective DQN-agent referred to as DQN-$i$. The resultant output is matching demonstration trajectory. It must be noticed that the resultant trajectory can be either a single demonstration or a sequential-combination of multiple demonstrations. 

The Higher-Level outputs the trajectory for the $N$-robots in the scene as $\hat{\mathcal{T}} := \{\boldsymbol{\hat{\mathcal{X}}}^{ee1}, \boldsymbol{\hat{\mathcal{X}}}^{ee2}, \cdots, \boldsymbol{\hat{\mathcal{X}}}^{eeN}\}$. This implies that for the $r^{th}$ robot the resultant trajectory is defined as, $\boldsymbol{\hat{\mathcal{X}}}^{eer} := \{\boldsymbol{x}^{ee}_0, \boldsymbol{x}^{ee}_1, \cdots, \boldsymbol{x}^{ee}_{Nr}\}$ where $N_r$ are the number of points in the resulting trajectory.

\subsection{ONCol-DMP: Lower Level Trajectory Execution}\label{sec:low_level}
\subsubsection{Optimized-Normalized DMP}
% XT modified 09/14/2024

The volume dynamic potential field \cite{Ginesi2021, Ginesi8981552} is applied for online obstacle avoidance, which added an additional perturbation term $\phi(\bm{x}, \bm{v})$ \footnote{As a notational convenience, we drop the subscripts and superscripts on $\bm{x}$ for ease of presentation in this section.} to the potential field:
    \begin{equation}
        \tau \dot{z}=\alpha_z(\beta_z (g-x)-z)+f(s)+\phi(\bm{x}, \bm{v}).
    \end{equation}

The dynamic potential function for the perturbation term $\phi(\bm{x}, \bm{v})$,  whose magnitude decreases with the distance $\rVert \bm{x}-\bm{o} \lVert$ and angle $\theta$ while increases with the system velocity $\rVert \bm{v} \lVert$, is defined as follows:
\begin{equation}
    U_D(\bm{x}, \bm{v}) = 
\begin{cases} 
\lambda \left(-\cos \theta \right)^\beta \frac{\lVert \bm{v} \rVert}{C^\eta(\bm{x})} & \text{if } \theta \in \left[\frac{\pi}{2}, \pi\right], \\
0 & \text{if } \theta \in \left[0, \frac{\pi}{2}\right],
\end{cases}
\label{eq: perturbation}
\end{equation}

In Eq.~\ref{eq: perturbation},
$C(\bm{x})$ is an ellipsoid isopotential function which indicates the distance between the obstacle and system:
\begin{equation}
C(\bm{x}) = \left( \frac{x_1 - o_1}{\ell_1} \right)^2 + \left( \frac{x_2 - o_2}{\ell_2} \right)^2 + \left( \frac{x_3 - o_3}{\ell_3} \right)^2
\end{equation}
where $x_1$, $x_2$, $x_3$ and $o_1$, $o_2$, $o_3$ are the respective components of the system's position $\bm{x}$ and the obstacle's center position $\bm{o}$ in the Cartesian coordinate system; $l_1$, $l_2$, $l_3$ denote the radii of the three principal axes of the ellipsoidal obstacle.
$\theta$ is the angle between the current velocity $\bm{v}$ and the system's position $\bm{x}$ relative to the position $\bm{o}$ of the obstacle:
\begin{equation}
    \theta=\mathrm{arccos}\left(\frac{<\bm{x}-\bm{o},\bm{v}>}{\lVert\bm{x}-\bm{o}\rVert \lVert\bm{v}\rVert} \right)
\end{equation}

 $\lambda$, $\beta$, $\eta$ are positive constant gains required to be optimized.
To ensure these parameters are adaptable to varying scales of reference trajectories, the reference trajectory is first transformed into a normalized space in the first quadrant through scaling, translation, and rotation. The potential field is then applied in this normalized space for obstacle avoidance, after which the deviated trajectory is mapped back to the original space.
The normalization mapping operation, denoted as $\mathcal{N}$, and the rescaling mapping back operation, denoted as $\mathcal{R}$, are defined as follows:
\begin{equation}
\overline{\bm{x}}=\mathcal{N}(\bm{x})=\frac{1}{\alpha_x}\bm{R}{(\bm{x}-\bm{b})},
\end{equation}
\begin{equation}
    \bm{x}=\mathcal{R}(\overline{\bm{x}})=
    \alpha_x \bm{R}^{-1} \overline{\bm{x}}+\bm{b},
\end{equation}

where $\bm{x}$ and $\overline{\bm{x}}$ represent the original and normalized trajectory respectively, $\alpha_x$ is the scaling parameter, $\bm{b}$ is the bias vector, $\bm{R}$ is the rotation matrix.

The force term $\phi(\bm{x}, \bm{v})$ is the negative gradient of the dynamic potential function $U_D(\bm{x}, \bm{v})$:
% \begin{equation}
% \begin{aligned}
\begin{multline}
\phi(\bm{x}, \bm{v}) = -\nabla_x (U_D(\bm{x}, \bm{v})) \\
 = -\nabla_x \left( \lambda (-\cos \theta)^\beta \frac{\lVert \bm{v} \rVert}{C^\eta(\bm{x})} \right) \\
 = \lambda \lVert \bm{v} \rVert (-\cos \theta)^{\beta-1} \left( -\beta \nabla_x (\cos \theta) + \frac{\eta \cos \theta}{C(\bm{x})} \nabla_x (C(\bm{x})) \right)
\end{multline}

The constrained objective function $f_c$ is designed to determine the parameters of potential field, which minimizes the energy consumption and the deviation between adaptive DMP trajectory $\bm{x_{a}}$ and reference trajectory $\bm{x_{r}}$ while avoiding obstacles:
% \begin{equation}
% \begin{aligned}
\begin{multline}\label{eq: cost func}
 f_c(\bm{p})=\sum_{J=1}^{N_t}\lVert\bm{x_{a}}(\bm{p},t_J)-\bm{x_{r}}(t_J)\rVert^2 \delta t \\
 + \frac{\lambda_p m}{2}  \sum_{J=1}^{N_t-1} |\lVert\bm{\dot{x}_{a}}(\bm{p},t_{J+1})\rVert^2 -\lVert\bm{\dot{x}_{a}}(\bm{p},t_J)\rVert^2 |\delta t,\\
\boldsymbol{s.t.}\ f_{cc}(\bm{p},t_J)=1-C(\bm{x_{a}}(\bm{p},t_J))<0, \forall \; J \in \{1,2,..., N_t\}   
\end{multline}
% \end{aligned}
% \label{eq: cost func}
% \end{equation}
where $t$ is the time, $T_{\mathrm{end}}$\footnote{We drop the subscript and use $T$ for rest of the manuscript.} is the period of the trajectory, $\bm{p}$ is the vector consists of the potential field parameters, i.e., $\bm{p}=[\lambda,\beta, \eta]$.
The first term in the Eq.~\ref{eq: cost func} quantifies the deviation between the obstacle avoidance and reference trajectories, while the second term in the Eq.~\ref{eq: cost func} measures energy consumption due to kinetic energy changes. Since both trajectories have identical initial point and goal, changes in gravitational potential energy are consistent, thus only kinetic energy is considered.
Once end-effector trajectory is determined by the DMP, we use inverse-kinematic solvers to compute the respective joint positions in real-time.

\subsubsection{Collaborative Execution}\label{sec:alpha_imp}
While the obstacle avoidance is accomplished using the potential field, it is done so by implementing it for each DMP independently. Since the DMP is only for the end effector, it fails to prevent collision between links of the two arms. To prevent such collisions and to improve the cooperation among the arms, we introduce a further improvisation to the DMP. Since, in some cases, one of the robot arm might obstruct the way or deviation from the trajectory might not feasible owing to the joint limits, we can choose to slow down the execution of one of the arm's DMP. To control the speed of execution of the DMP, we redefine the first order model of the phase transition from \label{eq:phase} as:
\begin{equation}
    \tau \dot{s} = -\alpha_s(\boldsymbol{x},\boldsymbol{x'})s
\end{equation}

where $\boldsymbol{x}$ and $\boldsymbol{x}'$ are two independent DMP variables. We compute $\alpha_s$ as:
\begin{equation}
    \alpha_s(\boldsymbol{x},\boldsymbol{x'}) = \hat{\alpha}(1-e^{-d})
\end{equation}
In the above equation $\hat{\alpha}$ is a constant parameter and can be same as defined in Sec.\ref{sec:background_dmp}. The variable $d$ is introduced as a measure of vicinity of the DMP trajectories at a given point and is defined as:
\begin{equation}
    d = 
\begin{cases}
    ||\boldsymbol{x} - \boldsymbol{x}'||_2,&  \text{if } ||\boldsymbol{x} - \boldsymbol{x}'||_2 \leq \epsilon_s\\
    \infty,              & \text{otherwise}
\end{cases}
\end{equation}

$\epsilon_s$ is a user defined parameter computed based on the dimensions of the end-effector to prevent collision similar to an inflation radius. Since the relaxation of the execution is imparted using the parameter $\alpha_s$ the exponential stability of the trajectory is remains intact.

\section{EXPERIMENTS AND VALIDATION}\label{sec:exper}
\begin{figure}[ht]
    \centering
    \includegraphics[width=\linewidth]{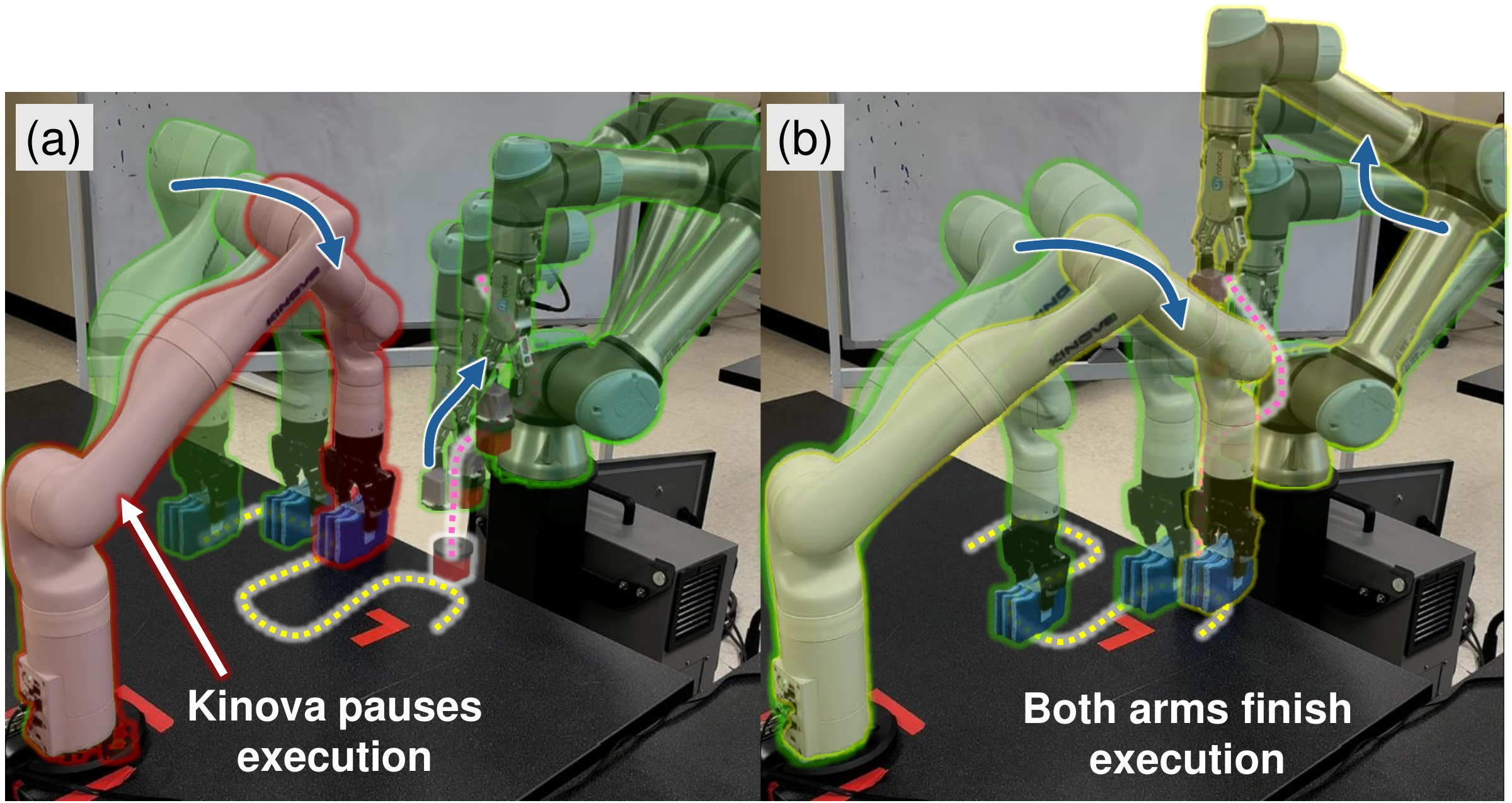}
    \caption{Hardware Implementation of ONCol DMP on a UR5e and Kinova Gen-3 robotic arm.}
    \label{fig:hardware}
\end{figure}

To validate the effectiveness of the proposed method we designed multiple experiments in simulation and on hardware. For the simulation purpose we use two UR5e robotic manipulators with Robotiq 2f-85 for end-effector on each arm. We refer to them as arm-1 and arm-2 for the remainder of the section. The simulation experimental setup is based in the PyBullet simulation environment. For the hardware setup we use a UR5e (6 DoF) manipulator with OnRobot RG2 gripper and a Kinova Gen-3 (7 DoF) manipulator with Robotiq 2f-85 gripper as shown in Fig.~\ref{fig:hardware}. 

We study the ability of the proposed method to i) devise online collision-free trajectories, ii) prevent collision amongst the arms and iii) complete distinct tasks collaboratively. For the purpose of simplicity we assume complete observability, i.e. the position and the velocity of each element in the scene is known and can be measured in real time.

For training the DQN agents we collected 10 different demonstration trajectories. The agents were trained using a RTX 3090 GPU with a AMD Ryzen Threadripper processor. Additionally, the DMP is executed at a frequency of 100 Hz i.e. each time step is 0.01 s. The parameters of the DMP are consistent for all tasks as follows:
$\hat{\alpha}=25/3, \alpha_z = 25, \beta_z=25/4.$
\begin{figure*}[ht]
      \centering
    \includegraphics[width=\linewidth]{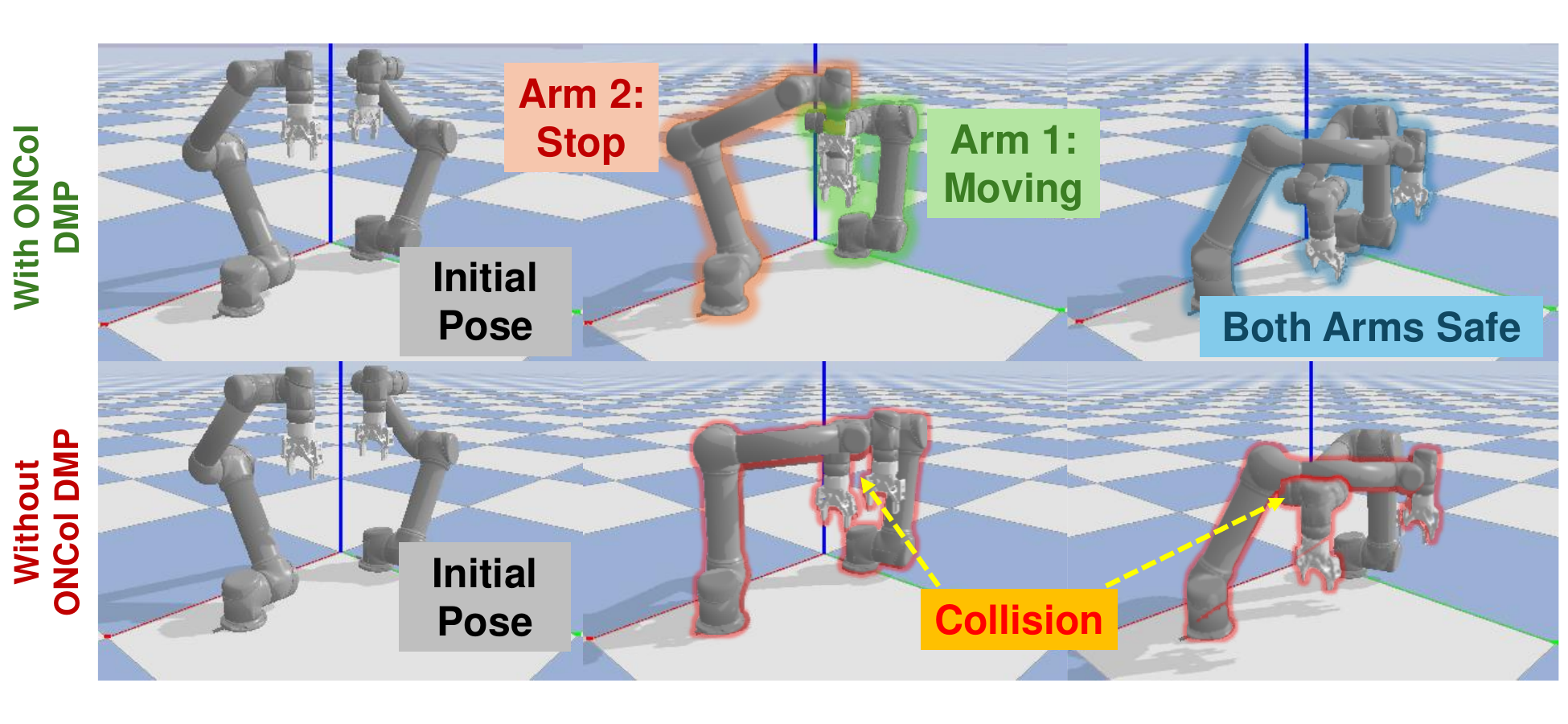}
    \caption{Setup in PyBullet environment for crossing arms. Top row shows the trajectory for the case with the proposed ONColDMP, whereas bottom row shows traditional DMP without collaborative phase control.}
    \label{fig:alpha_comp_pyb}
\end{figure*}

\begin{figure}[ht]
      \centering
    \includegraphics[width=\linewidth]{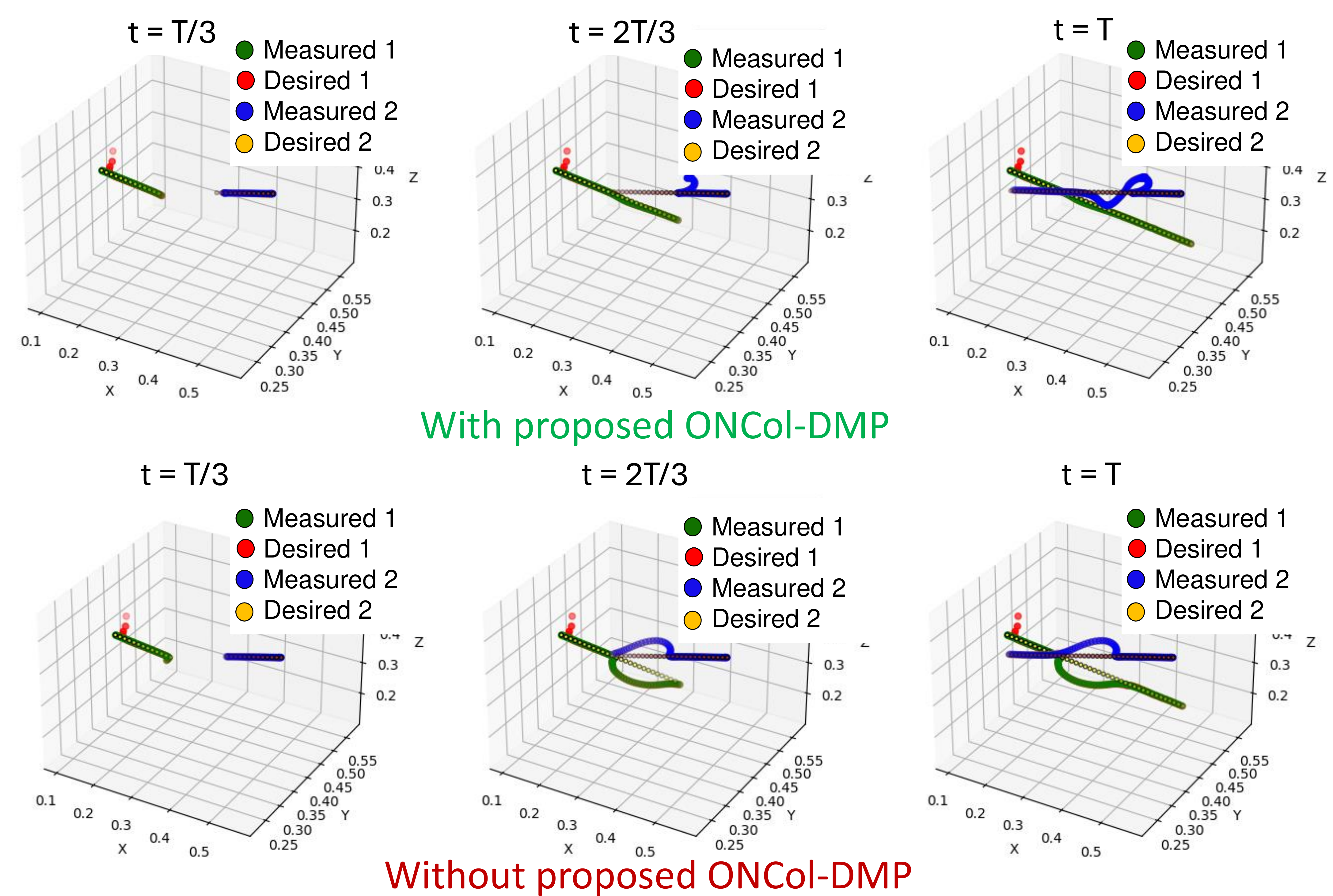}
    \caption{Comparison of the DMP trajectories in cartesian space for end-effector with the improvised ONCol DMP (top row) v/s without collaborative phase control term (bottom row).}
    \label{fig:alpha_comp}
\end{figure}

%I want to change this subtitle
% \subsection{Dynamic Obstacle Avoidance with ONDMP}
% The proposed changes to the DMP formulation allows us to navigate around mobile obstacles of varying sizes. To characterize the efficacy of the proposed method we conduct a simple obstacle avoidance scenario. Fig. \ref{fig:speed_vs_avoid} showcases the deviation when three different blocks are moved along the y-axis while the robot end-effector must translate along the x-axis. As it can be seen from the plots, the deviation is maximum for 0.1 m/s and is generally higher at faster speeds compared to lower speeds. This is owing to the fact that at faster speeds the obstacle moves out of the faster as well hence requiring less avoidant behavior from the DMP. 
% \begin{figure}
%     \centering
%     \includegraphics[width=1\linewidth]{Dev_vs_speed.png}
%     \caption{Enter Caption}
%     \label{fig:speed_vs_avoid}
% \end{figure}
\begin{figure}[ht]
      \centering
    \includegraphics[width=\linewidth]{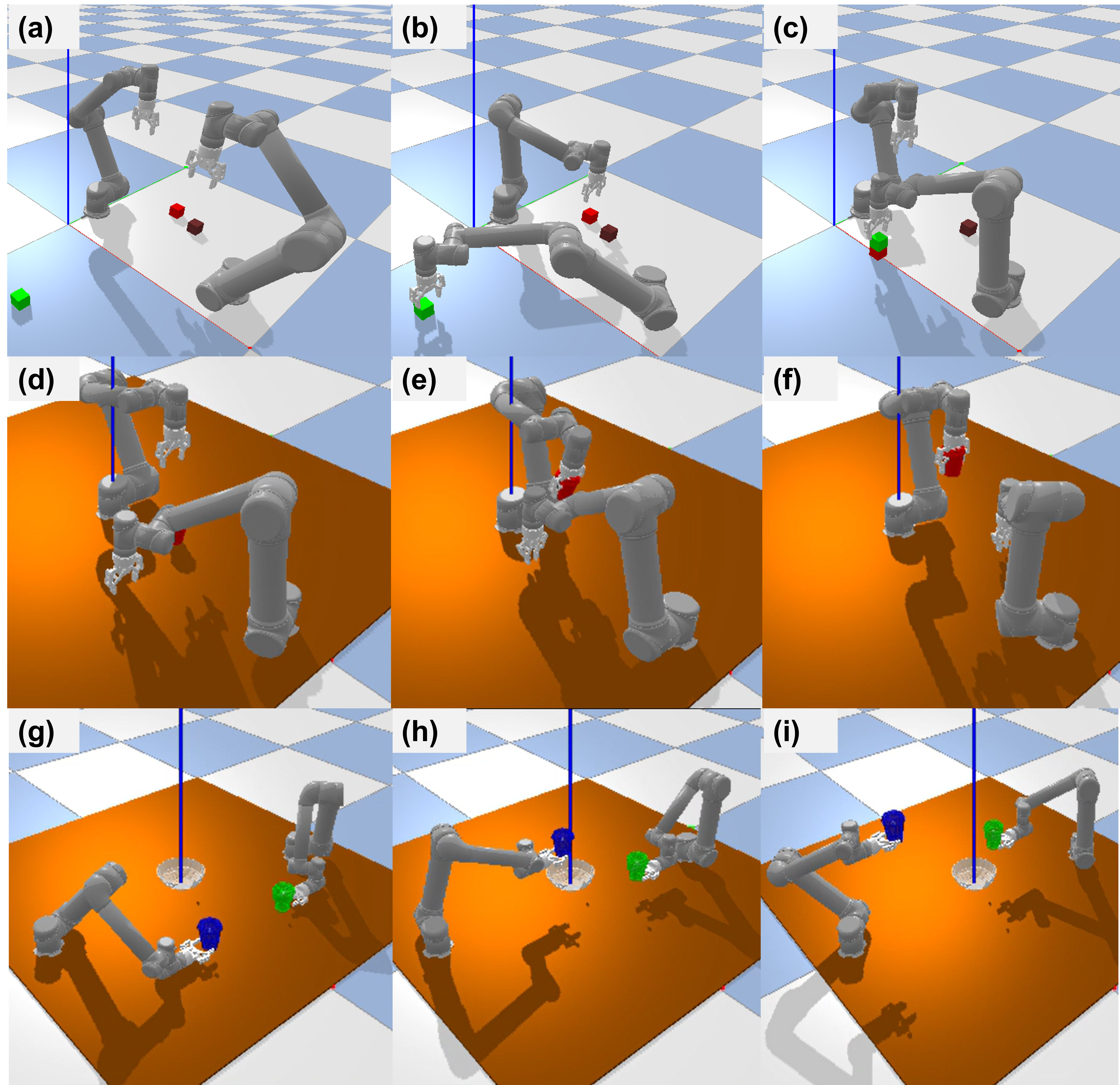}
    \caption{PyBullet setup showing collaborative with two arms.(a-c) Block stacking, (d-f) Table cleaning, and (g-i) water transfer.}
    \label{fig:scenario_pyb}
\end{figure}

\subsection{Cross-Trajectory Collaborative Task}
We study the improvement of the trajectory execution as proposed in Sec.\ref{sec:alpha_imp}. To better highlight the strength of the method we purposely devise trajectories with starting and ending goal positions which require the two arms to cross each other. We simulate the experiment in the PyBullet environment as shown in Fig.~\ref{fig:alpha_comp_pyb}. As indicated in Fig.~\ref{fig:alpha_comp}, introducing the collaborative behavior reduces the execution speed of arm-2 while the arm-1 continues to execute its trajectory. This leads to a much smaller deviation in the trajectory as compared to a case where the other arm is only considered as an obstacle. For the purpose of this experiment $\hat{\alpha} = 25/3$ and $r_s = 0.25$. All the other parameters were the same for the two trajectories. Table \ref{tab:dev_comp} shows the comparison of performance for the two cases. We observe the maximum deviation a parameter of performance. It is evident that utilizing the collaborative term reduces the deviation in the motion by introducing a sequential-like approach. Since, only the second robot was equipped with the enhanced $\alpha_s$, the reduced deviation for arm-1 coincides with the hypothesis.

\begin{table}
    \centering
    \caption{Max. Deviation of Arms in Crossing Task}
    \begin{tabular}{c|c|c}
        \hline
             & \textbf{Arm-1} & \textbf{Arm-2} \\
            \hline \hline
         \textbf{DMP} & 0.12 m & 0.14 m \\
        \textbf{ONCol DMP} &  0.01 m & 0.13 m\\
        \hline
    \end{tabular}
    \label{tab:dev_comp}
\end{table}
\begin{figure}[ht]
      \centering
    \includegraphics[width=\linewidth]{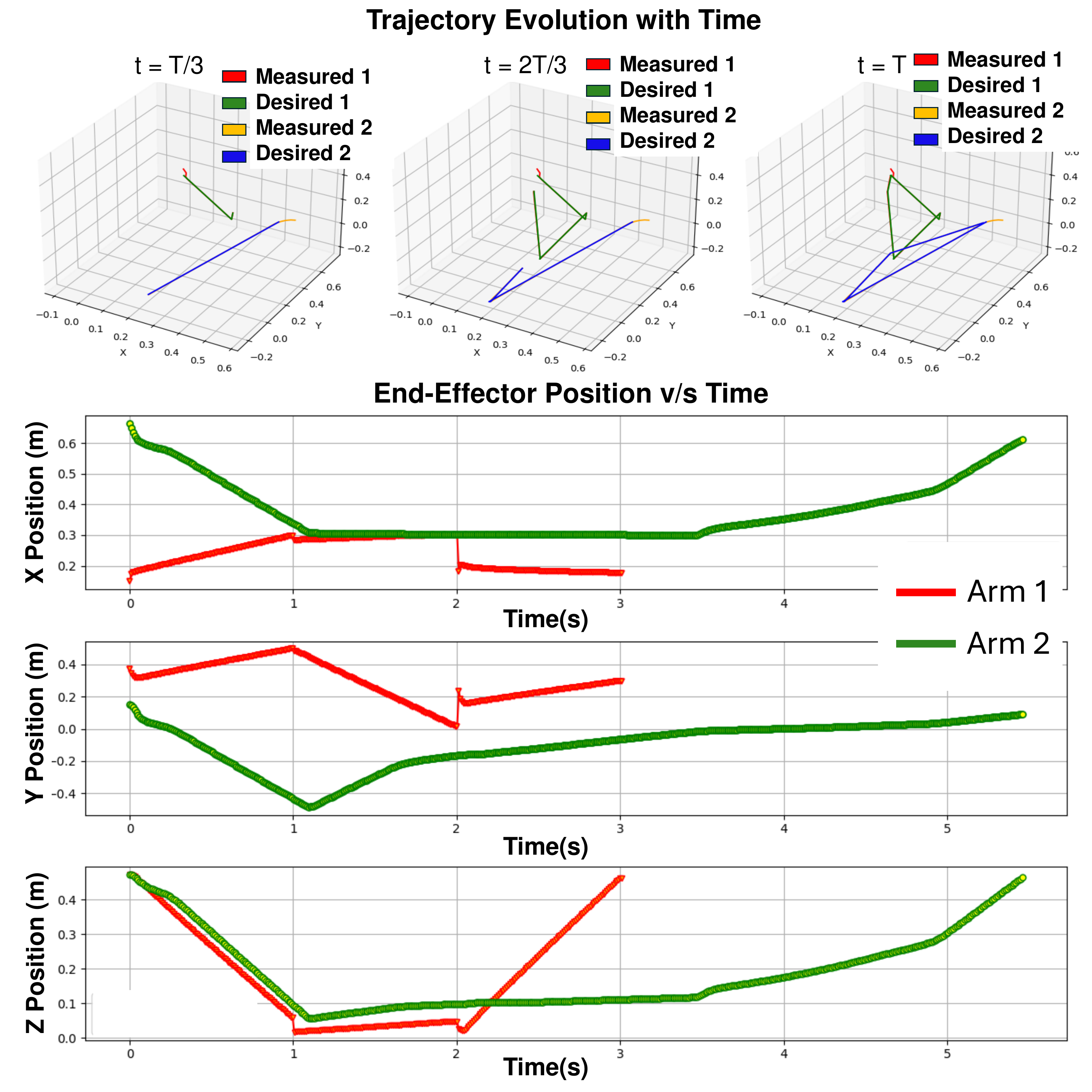}
    \caption{Trajectory evolution of the two end-effectors with time. (Top row) The $(x,y,z)$ position of the end-effector after three equal intervals. (Bottom three rows) Plot of $(x,y,z)$ position $\; v/s \; time (s)$ for the two end-effectors.}
    \label{fig:stacking_traj}
\end{figure}

\subsection{Long Sequence Collaborative Tasks}
We further extend our method to demonstrate the ability to handle distinct long sequence tasks as shown in Fig.~\ref{fig:scenario_pyb}, namely block stacking Fig.~\ref{fig:scenario_pyb} (a-c), table cleaning Fig.~\ref{fig:scenario_pyb} (d-f), and water pouring Fig.~\ref{fig:scenario_pyb} (g-i). The task constraints are provided as a sequence of critical-configurations as discussed in Sec.~\ref{sec:high_level}. The proposed method lead to successful collision free completion of task for all three cases. 
Due to reason of space constraints we showcase the behavior of the block stacking task \footnote{For videos and more details of different tasks please visit the project website: \url{https://sites.google.com/virginia.edu/oncoldmp/home}}. Figure~\ref{fig:stacking_traj} shows the trajectory for the block stacking task involving two UR5e arms. Since the stack has to be made at a single place, when arm-1 places a block, it can act as an obstacle for arm-2. However, as evident from the plots, the motion along the y-axis, starts slowing down for arm-2 after the 1.6 s mark, whereas the arm-1 continues the motion at the desired speed. Once arm-1 places the block, it moves out of the way and arm-2 can continue placing block. The delay causes the arm-2 DMP to be executed for longer duration. 

We further validated our technique on hardware setup as shown in Fig.~\ref{fig:hardware} using Kinova 7-DoF arm and a UR5e 6-DoF arm further highlighting the adaptability of the proposed method. 

% \subsection{Validation on Physical Setup}
% In real-world physical experiments, a UR5e (6-DoF) robot and a Kinova Gen-3 (7-DoF) robotic arm are used, as shown in Figure\ref{fig:hardware}. The UR5e is tasked with simply picking up an object from a table which happens to lie on the siping trajectory that the Kinova must follow. As shown in the Fig. \ref{fig:hardware}(a), the Kinova arm wipes the table but gradually slows down and stops as the UR5e approaches the block. The UR5e then picks up the block while avoiding the Kinova’s end-effector. Once the UR5e lifts the block and leaves enough safe distance for the Kinova’s execution, the Kinova manipulator gradually resumes wiping the table and both the UR5e and Kinova complete their respective tasks as shown in Fig. \ref{fig:hardware}(b).
% \begin{figure}[t!]
%     \centering
%     \begin{subfigure}[t]{\linewidth}
%         \centering
%         \includegraphics[width=0.5\linewidth]{Hardware1.pdf}
%         \caption{Kinova pausing the execution of the trajectory to prevent collision with Ur5e.}
%     \end{subfigure}
%     \begin{subfigure}[t]{\linewidth}
%         \centering
%         \includegraphics[width=0.5\linewidth]{Hardware2.pdf}
%         \caption{Kinova continues the reference trajectory tracking after the Ur5e moves to a feasible region.}
%     \end{subfigure}
%     \caption{Hardware Implementation of ONCol DMP ona UR5e and Kinova Gen-3 robotic arm.}
% \end{figure}

% \begin{figure}
%     \centering
%     \includegraphics[width=1\linewidth]{Hardware_alt.png}
%     \caption{Enter Caption}
%     \label{fig:enter-label}
% \end{figure}

\subsection{Multi-Arm Scenario}
\begin{figure}[ht]
      \centering
    \includegraphics[width=\linewidth]{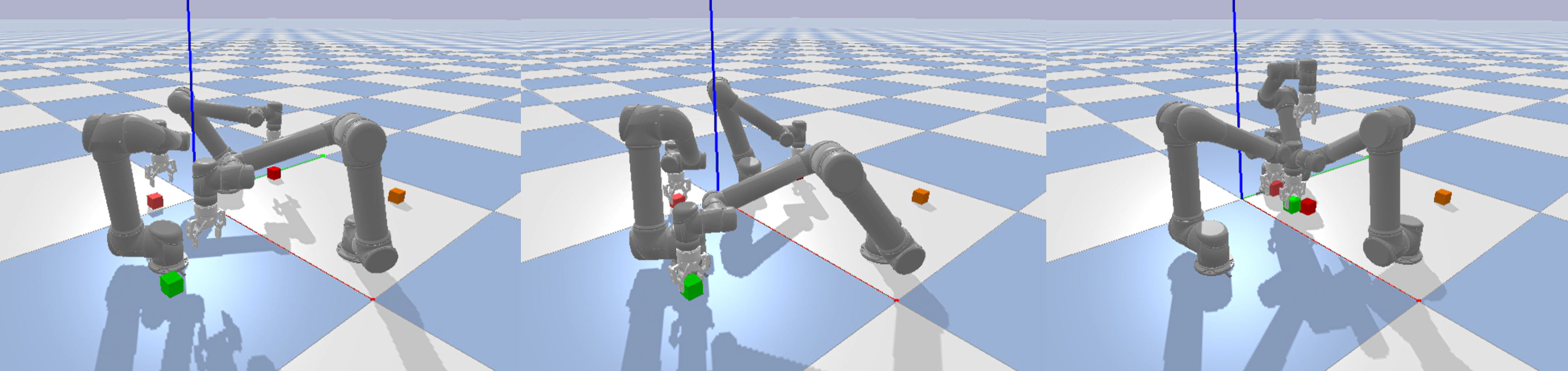}
    \caption{PyBullet setup showing collaborative task of stacking with three UR5e arms.}
    \label{fig:mult_arm}
\end{figure}

Since the proposed method is not limited by the number of agents and the training of LfD agents is independent, this allows us to extend the method without any changes to the larger number of arms. We validate this by extending our method to a multi-arm setup as shown in Fig.~\ref{fig:mult_arm} for a block stacking task. 

\section{CONCLUSIONS AND FUTURE WORK}\label{sec:conclusion}
In this work we proposed a novel hierarchical method which leveraged human demonstrations based RL technique to generate a motion plan at higher level and employed an improvised DMP to foster collaborative effort for multi-arm systems.

Through various experiments we identified that the proposed method can: i) generate trajectories for multiple arms given their respective task constraints, ii) avoid obstacles dynamically without parameter re-tuning, and iii) accomplish collaborative tasks by heuristic collaborative phase control. The proposed technique can be deployed to generate new trajectories online and in real time. Users can enhance the ability of the robots to generate contextual trajectories by either adding more demonstrations or modifying the existing ones. For example, in a robotic assembly task the  demonstrated skills can specifically focus on screwing/unscrewing, bolting, stacking etc., giving the users a more flexible setup. The proposed method works as an off-the-shelf plug and play tool since it only requires training a single RL agent and then deploying to multiple arms without re-training facilitating collaboration. The independence of the trained RL agent also allows user to impart distinct behaviors/roles to different arms.  

While the proposed method successfully avoids collisions between the end-effector and the links using a combination of potential fields and phase control, the current heuristic approach cannot guarantee collision avoidance among the links and may lead to deadlocks if not deployed correctly. In the future, we aim to develop a multi-agent system to enhance collaboration and reduce the possibility of collisions among the robot links.


%%%%%%%%%%%%%%%%%%%%%%%%%%%%%%%%%%%%%%%%%%%%%%%%%%%%%%%%%%%%%%%%%%%%%%%%%%%%%%%%

%%%%%%%%%%%%%%%%%%%%%%%%%%%%%%%%%%%%%%%%%%%%%%%%%%%%%%%%%%%%%%%%%%%%%%%%%%%%%%%%



\section*{ACKNOWLEDGMENT}

This work is supported by the National Science Foundation Grant CMMI 2243930 and 1853454.

\bibliographystyle{IEEEtran}
\bibliography{main}

\begin{thebibliography}{10}
\providecommand{\url}[1]{#1}
\csname url@samestyle\endcsname
\providecommand{\newblock}{\relax}
\providecommand{\bibinfo}[2]{#2}
\providecommand{\BIBentrySTDinterwordspacing}{\spaceskip=0pt\relax}
\providecommand{\BIBentryALTinterwordstretchfactor}{4}
\providecommand{\BIBentryALTinterwordspacing}{\spaceskip=\fontdimen2\font plus
\BIBentryALTinterwordstretchfactor\fontdimen3\font minus \fontdimen4\font\relax}
\providecommand{\BIBforeignlanguage}[2]{{%
\expandafter\ifx\csname l@#1\endcsname\relax
\typeout{** WARNING: IEEEtran.bst: No hyphenation pattern has been}%
\typeout{** loaded for the language `#1'. Using the pattern for}%
\typeout{** the default language instead.}%
\else
\language=\csname l@#1\endcsname
\fi
#2}}
\providecommand{\BIBdecl}{\relax}
\BIBdecl

\bibitem{dualarmsurvey}
C.~Smith, Y.~Karayiannidis, L.~Nalpantidis, X.~Gratal, P.~Qi, D.~V. Dimarogonas, and D.~Kragic, ``Dual arm manipulation—a survey,'' \emph{Robot. Auton. Syst.}, vol.~60, no.~10, pp. 1340--1353, Oct. 2012.

\bibitem{schaal}
S.~Schaal, ``Is imitation learning the route to humanoid robots?'' \emph{Trends in Cognitive Sciences}, vol.~3, no.~6, pp. 233--242, Jun. 1999.

\bibitem{collab3}
A.~Tung and et~al., ``Learning multi-arm manipulation through collaborative teleoperation,'' in \emph{2021 IEEE International Conference on Robotics and Automation (ICRA)}, Xi'an, China, 2021, pp. 9212--9219.

\bibitem{RLSurvey}
J.~Kober, J.~A. Bagnell, and J.~Peters, ``Reinforcement learning in robotics: A survey,'' \emph{Int. J. Robot. Res.}, vol.~32, no.~11, pp. 1238--1274, Sep. 2013.

\bibitem{Levine}
S.~Levine, C.~Finn, T.~Darrell, and P.~Abbeel, ``End-to-end training of deep visuomotor policies,'' \emph{J. Mach. Learn. Res.}, vol.~17, no.~1, pp. 1334--1373, 2016.

\bibitem{Gl1}
S.~W. Sohn, K.~Oh, J.~Kang, and J.~H. Kim, ``Graph-based reinforcement learning for robotic task sequencing,'' \emph{IEEE Trans. Robot.}, vol.~37, no.~4, pp. 1108--1119, Aug. 2021.

\bibitem{Gl2}
F.~Yang, M.~Chen, D.~D. Lee, and H.~Zhang, ``Learning task sequencing for complex manipulation tasks with temporal constraints,'' in \emph{Proc. IEEE Int. Conf. Robot. Autom. (ICRA)}, 2021, pp. 8791--8797.

\bibitem{rulebased}
J.~K. Behrens, R.~Lange, and M.~Mansouri, ``A constraint programming approach to simultaneous task allocation and motion scheduling for industrial dual-arm manipulation tasks,'' in \emph{2019 International Conference on Robotics and Automation (ICRA)}, Montreal, QC, Canada, 2019, pp. 8705--8711.

\bibitem{logic1}
S.~Saha and A.~A. Julius, ``Task and motion planning for manipulator arms with metric temporal logic specifications,'' \emph{IEEE Robot. Autom. Lett.}, vol.~3, no.~1, pp. 379--386, Jan. 2018.

\bibitem{mpc1}
D.~Q. Mayne, J.~B. Rawlings, C.~V. Rao, and P.~O. Scokaert, ``Constrained model predictive control: Stability and optimality,'' \emph{Automatica}, vol.~36, no.~6, pp. 789--814, Jun. 2000.

\bibitem{mpc2}
X.~Zhao, Y.~Zhang, W.~Ding, B.~Tao, and H.~Ding, ``A dual-arm robot cooperation framework based on a nonlinear model predictive cooperative control,'' \emph{IEEE/ASME Trans. Mechatronics}, 2023.

\bibitem{to1}
Y.~Tassa, T.~Erez, and E.~Todorov, ``Synthesis and stabilization of complex behaviors through online trajectory optimization,'' in \emph{Proc. IEEE Int. Conf. Robot. Autom. (ICRA)}, 2012, pp. 4906--4913.

\bibitem{nadia1}
S.~S.~M. Salehian, N.~B.~F. Fernandez, and A.~Billard, ``Coordinated multi-arm motion planning: Reaching for moving objects in the face of uncertainty,'' in \emph{Proc. Robot. Sci. Syst. Conf.}, 2016.

\bibitem{nadia2}
S.~S.~M. Salehian and et~al., ``Dynamical system-based motion planning for multi-arm systems: Reaching for moving objects,'' \emph{Auton. Robot.}, vol.~40, no.~4, pp. 649--671, Apr. 2016.

\bibitem{Ginesi2021}
M.~Ginesi, D.~Meli, A.~Roberti, N.~Sansonetto, and P.~Fiorini, ``Dynamic movement primitives: Volumetric obstacle avoidance using dynamic potential functions,'' \emph{Auton. Robot.}, vol.~46, no.~2, pp. 173--196, Feb. 2022.

\bibitem{Ginesi8981552}
M.~Ginesi, D.~Meli, A.~Calanca, D.~Dall'Alba, N.~Sansonetto, and P.~Fiorini, ``Dynamic movement primitives: Volumetric obstacle avoidance,'' in \emph{2019 19th International Conference on Advanced Robotics (ICAR)}, Belo Horizonte, Brazil, 2019, pp. 234--239.

\bibitem{tianyu}
T.~Yu and Q.~Chang, ``Motion planning for human-robot collaboration based on reinforcement learning,'' in \emph{2022 IEEE 18th International Conference on Automation Science and Engineering (CASE)}, Mexico City, Mexico, 2022, pp. 1866--1871.

\bibitem{Atkeson1997LocallyWL}
C.~G. Atkeson, A.~W. Moore, and S.~Schaal, ``Locally weighted learning,'' \emph{Artif. Intell. Rev.}, vol.~11, no. 1--5, pp. 11--73, Apr. 1997.

\end{thebibliography}

\end{document}